\documentclass[11pt,a4paper]{article}
\usepackage[hyperref]{emnlp2020}
\usepackage{times}
\usepackage{latexsym}
\usepackage{subfig}
\usepackage{hyperref}

\usepackage[hang,flushmargin]{footmisc}  % minimize footnote indentation
\usepackage{amsfonts,amsmath,amssymb}
\usepackage[T1]{fontenc}  % fix font related glitches
\usepackage{bold-extra}   % enable \texttt{\textbf{}}
\usepackage{microtype}    % make margins prettier
\usepackage{graphicx}
\usepackage{multirow}
\usepackage{bm}           % bold font in math mode
\usepackage{soul}         % highlight
\usepackage{algorithm}
\usepackage[noend]{algpseudocode}
\usepackage{array,booktabs}
\usepackage{stfloats}  % for bottom-aligned table

\usepackage{enumitem}
\setenumerate[1]{leftmargin=*}    % no enumerate indentation
\setitemize[1]{leftmargin=*}      % no itemize indentation

\aclfinalcopy % Uncomment this line for the final submission
\newcommand{\textsec}[1]{\textsection\ref{#1}}
\newcommand\LN{\linebreak\noindent}
\newcommand\AAA{\texttt{AA}}  % \AA already defined
\newcommand\EE{\texttt{EE}}
\newcommand\SC{\texttt{SC}}
\newcommand\CM{\texttt{CM}}

\title{Revealing the Myth of Higher-Order Inference in Coreference Resolution}

\author{Liyan Xu \\
  Computer Science \\
  Emory University, Atlanta, GA \\
  \texttt{liyan.xu@emory.edu} \\\And
  Jinho D. Choi \\
  Computer Science \\
  Emory University, Atlanta, GA \\
  \texttt{jinho.choi@emory.edu} \\}

\date{}

\begin{document}
\maketitle

\begin{abstract}
This paper analyzes the impact of higher-order inference (HOI) on the task of coreference resolution.
HOI has been adapted by almost all recent coreference resolution models without taking much investigation on its true effectiveness over representation learning.
To make a\LN comprehensive analysis, we implement an end-to-end coreference system as well as four HOI approaches, attended antecedent, entity equalization, span clustering, and cluster merging, where the latter two are our original methods.
We find that given a high-performing encoder such as SpanBERT, the impact of HOI is negative to marginal, providing a new perspective of HOI to this task.
Our best model using cluster merging shows the Avg-F1 of 80.2 on the CoNLL 2012 shared task dataset in English.
%, that is the new state-of-the-art result for this task.
\end{abstract}

\section{Introduction}
\label{sec:intro}

\begin{figure}[htbp!]
\centering
\includegraphics[width=0.9\columnwidth]{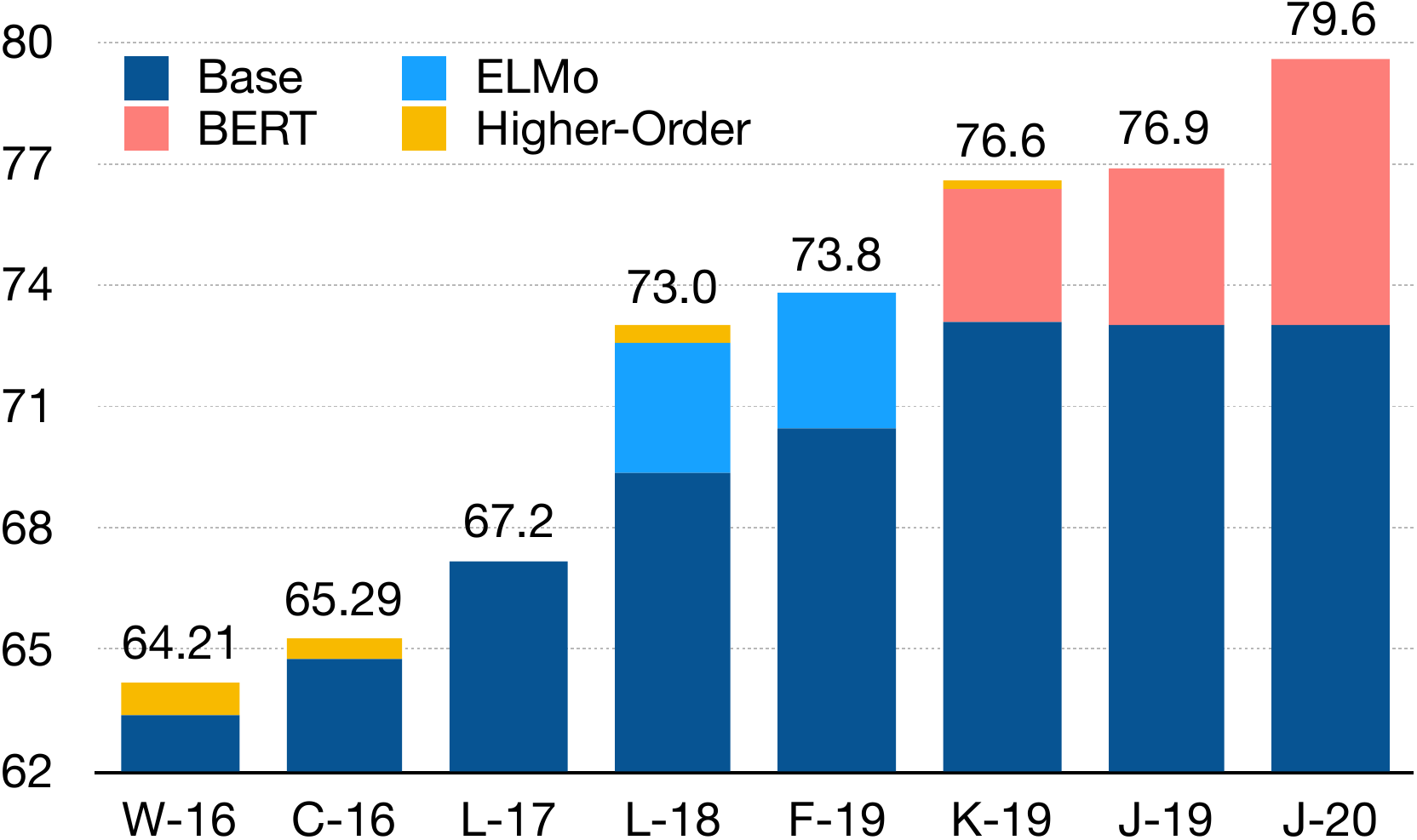}
\caption{Performance breakdown of the recent state-of-the-art models on the CoNLL 2012 shared task. W-16: \citet{wiseman-etal-2016-learning}, C-16: \citet{clark-manning-2016-improving}, L-17: \citet{lee-etal-2017-end}, L-18: \citet{lee-etal-2018-higher}, F-19: \citet{fei-etal-2019-end}, K-19: \citet{kantor-globerson-2019-coreference}, J-19: \citet{joshi-etal-2019-bert}, J-20: \citet{spanbert-joshi}.}
\label{fig:coref-history}
\end{figure}

Coreference resolution has always been considered one of the unsolved NLP tasks due to its challenging aspect of document-level understanding \cite{wiseman-etal-2015-learning,wiseman-etal-2016-learning,clark-manning-2015-entity,clark-manning-2016-improving,lee-etal-2017-end}.
Nonetheless, it has made a tremendous progress in recent years by adapting contextualized embedding encoders such as ELMo \cite{lee-etal-2018-higher,fei-etal-2019-end} and BERT \cite{kantor-globerson-2019-coreference,joshi-etal-2019-bert,spanbert-joshi}.
The latest state-of-the-art model shows the improvement of 12.4\% over the model introduced 2.5 years ago, where the major portion of the improvement is derived by representation learning (Figure~\ref{fig:coref-history}).

Most of these previous models have also adapted higher-order inference (HOI) for the global optimization of coreference links, although HOI clearly has not been the focus of those works, for the fact that gains from HOI have been reported marginal.
This has inspired us to analyze the impact of HOI on modern coreference resolution models in order to envision the future direction of this research.

\noindent To make thorough ablation studies among different approaches, we implement an end-to-end coreference system in PyTorch (Sec~\ref{subsec:baseline}),
% that can adapt any contextualized embedding encoder (Section~\ref{subsec:baseline})
and two HOI approaches proposed by previous work, attended antecedent and entity equalization (Sec~\ref{subsec:previous}), along with two of our original approaches, span clustering and cluster merging (Sec~\ref{subsec:new}).
These approaches are experimented with two Transformer encoders, BERT and SpanBERT, to assess how effective HOI is even when coupled with those high-performing encoders (Sec~\ref{sec:exp}).
To the best of our knowledge,\LN this is the first work to make a comprehensive analysis on multiple HOI approaches side-by-side for the task of coreference resolution.\footnote{Source codes and models are available at \url{https://github.com/lxucs/coref-hoi}.}

\vspace{-1.5ex}
\section{Related Work}
\label{sec:related}

Most neural network-based coreference resolution models have adapted antecedent-ranking \cite{wiseman-etal-2015-learning,clark-manning-2015-entity,lee-etal-2017-end,lee-etal-2018-higher,joshi-etal-2019-bert,spanbert-joshi}, which relies on the local decisions between each mention and its antecedents.
To achieve deeper global optimization,\LN \citet{wiseman-etal-2016-learning,clark-manning-2016-improving,yu-etal-2020-cluster} built entity representations in the ranking process, whereas \citet{lee-etal-2018-higher,kantor-globerson-2019-coreference} refined the mention representation by aggregating its antecedents' information.

It is no secret that the integration of contextualized embeddings has played the most critical role in this task. While the following are based on the same end-to-end coreference model \cite{lee-etal-2017-end},
\citet{lee-etal-2018-higher,fei-etal-2019-end} reported 3.3\% improvement by adapting ELMo in the encoders \cite{peters-etal-2018-deep}.
% \citet{lee-etal-2018-higher,fei-etal-2019-end} adapted ELMo \cite{peters-etal-2018-deep} that gave 3.3\% improvement over the previous state-of-the-art model \cite{lee-etal-2017-end}.
\citet{kantor-globerson-2019-coreference,joshi-etal-2019-bert} gained additional 3.3\% by adapting BERT \cite{devlin-etal-2019-bert}.
\citet{spanbert-joshi} introduced SpanBERT that gave another 2.7\% improvement over \citet{joshi-etal-2019-bert}.

Most recently, \citet{wu-etal-2020-corefqa} proposes a new model that adapts question-answering framework on coreference resolution, and achieves state-of-the-art result of 83.1 on the CoNLL'12 shared task.

%The antecedent-ranking-based coreference models have been popular in recent years because of its simplicity and superior empirical results. The model learns to predict one single antecedent for each anaphoric mention \citep{wiseman-etal-2015-learning,wiseman-etal-2016-learning,clark-manning-2016-improving,lee-etal-2017-end,lee-etal-2018-higher}. More recently, the use of pretrained Transformers such as BERT \citep{devlin-etal-2019-bert} and SpanBERT \citep{spanbert-joshi} on the representation level has achieved significant performance gain for the coreference models.
%While the vanilla antecedent-ranking only involves local decisions between each pair of the span and its antecedent, different higher-order inference techniques have been proposed. \citet{lee-etal-2018-higher} and ; 

\section{Approach}
\label{sec:approach}

%Four different HOI approaches are experimented on the same baseline model (Section~\ref{subsec:baseline}). We refer to the two previous approaches as Attended Antecedent (\AAA), Entity Equalization (\EE) (Section~\ref{subsec:previous}), and two new approaches as Span Clustering (\SC), Cluster Merging (\CM) (Section~\ref{subsec:new}).

\subsection{End-to-End Coreference System}
\label{subsec:baseline}

We reimplement the end-to-end \textit{c2f-coref} model introduced by \citet{lee-etal-2018-higher} that has been adapted by every coreference resolution model since then. It detects mention candidates through span enumeration and aggressive pruning.
For each candidate span $x$, the model learns the distribution over its antecedents $y \in \mathcal{Y}(x)$:
\begin{equation}
    P(y) = \frac{e^{s(x, y)}}{\sum_{y' \in \mathcal{Y}(x)}e^{s(x, y')}} \label{eq:py}
\end{equation}
where $s(x,y)$ is the local score involving two parts: how likely the spans $x$ and $y$ are valid mentions, and how likely they refer to the same entity:
\begin{align}
\label{eq:antecedent_score}
    s(x, y)  &= s_m(x) + s_m(y) + s_c(x,y)\\
    s_m(x)   &= w_m \texttt{FFNN}_m (g_x) \nonumber\\
    s_c(x,y) &= w_c \texttt{FFNN}_c (g_x, g_y, \phi(x,y)) \nonumber
\end{align}
$g_x, g_y$ are the span embeddings of $x$ and $y$, $\phi(x,y)$ is the meta-information (e.g., speakers, distance), and $w_m, w_c$ are the mention and coreference scores, respectively (\texttt{FFNN}: feedforward neural network).

We use different Transformers-based encoders, and follow the ``\textit{independent}'' setup for long documents as suggested by \citet{joshi-etal-2019-bert}.

%This system can adapt any embedding encoder as well as HOI strategies in the following sections, which makes it easy to compare different methods fairly and analyze them with fine-grained control.

%We use the state-of-the-art model Transformers-based \textit{c2f-coref} as the baseline \citep{joshi-etal-2019-bert,spanbert-joshi}. The encoding part is the Pretrained Transformers which is finetuned during training;
% We follow the same \textit{independent} variant as suggested, where non-overlapping segments are fed to Transformers, acting as independent instances of a document.
% and the span representation is the concatenation of the first and last word piece as well as all attended word pieces in the span.

\subsection{Span Refinement}
\label{subsec:previous}

Two HOI methods presented by recent coreference work are based on span refinement that aggregates non-local features to enrich the span representation with more ``global'' information.
The updated span representation $g_x'$ can be derived as in Eq.~\ref{eq:refine}, where $g_x'$ is the interpolation between the current and refined representation $g_x$ and $a_x$, and $W_f$ is the gate parameter. 
$g_x'$ is used to perform another round of antecedent-ranking in replacement of $g_x$.
\begin{align}
\label{eq:refine}
    g_x' &= f_x \circ g_x + (1 - f_x) \circ a_x \\
    f_x &= \sigma (W_f [g_x, a_x])\nonumber
\end{align}
The following two methods share the same updating process for $g_x'$, but with different ways to obtain the refined span representation $a_x$.

\paragraph{Attended Antecedent (\texttt{AA})} 

takes the antecedent information to enrich $g_x'$ \citep{lee-etal-2018-higher,fei-etal-2019-end,joshi-etal-2019-bert,spanbert-joshi}. 
The refined span $a_x$ is the attended antecedent representation over the current antecedent distribution $P(y)$, where $g_{y \in \mathcal{Y}(x)}$ is the antecedent representation:
\begin{equation}
    a_x = \sum_{y \in \mathcal{Y}(x)} P(y) \cdot g_y
\end{equation}

\paragraph{Entity Equalization (\texttt{EE})}

takes the clustering relaxation as in Eq.~\ref{eq:ee} to model the entity distribution \citep{kantor-globerson-2019-coreference}, where $Q(x \in E_{y'})$ is the probability of the span $x$ referring to an entity $E_{y'}$ in which the span $y'$ is the first mention. $P(y)$ is the current antecedent distribution.
\begin{align}
\label{eq:ee}
    & Q(x \in E_{y'}) = \nonumber\\
    & \begin{cases}
        \sum_{k=y'}^{x-1} P(y = k) \cdot Q(k \in E_{y'}) & y' < x \\
        P(y = \epsilon) & y' = x \\
        0 & y' > x
    \end{cases}
\end{align}
The refined span $a_x$ is the attended entity representation, where $e_y^{(x)}$ is the entity representation to which the span $y$ belongs till the span $x$:
\begin{align}
    e_x^{(t)} &= \sum_{y=1}^t Q(y \in E_x) \cdot g_y \\
    a_x &= \sum_{y=1}^x Q(x \in E_y) \cdot e_y^{(x)}
\end{align}

%texttt{EE} uses the more ``global'' entity-level information.

\subsection{HOI with Clustering}
\label{subsec:new}

This section introduces two new HOI methods for a more extensive study in HOI.
%In particular, \texttt{SC} is also based on span refinement, while \texttt{CM} is similar to the idea of \citet{wiseman-etal-2016-learning} that directly maintains the entity clusters during antecedent-ranking.

\paragraph{Span Clustering (\texttt{SC})}

is also based on span refinement, and it
constructs the actual clusters and obtains the ``true'' predicted entities using $P(y)$ instead of modeling the ``soft'' entity clusters through the relaxation as in \texttt{EE} (Section~\ref{subsec:previous}).
This way, although we lose the differentiable property, the obtaining of true entities with the same empirical inference time as \texttt{EE} has made \texttt{SC} desirable.

\noindent The entity representation $e_i$ for an entity cluster $C_i$ is given by the attended spans in this cluster:
\begin{align*}
    \alpha_t &= w_{\alpha} \text{FFNN}_{\alpha} (g_t) \\
    \alpha_{i,t} &= \frac{\exp(\alpha_t)}{\sum_{k \in C_i}\exp(\alpha_k)} \\
    e_i &= \sum_{t \in C_i} \alpha_{i,t} \cdot g_t
\end{align*}
The entity clusters $C_i$ are constructed in the same way as in the final cluster prediction. 
The refined span $a_x$ is then equal to the representation of entity $e_i$ to which it belongs ($g_x \in C_i$).

\paragraph{Cluster Merging (\texttt{CM})}

performs sequential antecedent ranking combining both antecedent and entity information to gradually build up the entity clusters, which is distinguished from span refinement methods that simply re-rank antecedents.
Algorithm~\ref{algo:cm} describes the ranking process for \texttt{CM}. 
$g_i$ is the $i$'th span, $\mathcal{Y}(i)$ is the indices of $g_i$'s antecedents, and $C_i$ is the cluster that $g_i$ belongs to.
The ranking score  $s_x(y)$ consists of both antecedent score $f_a$ (see Eq.~\ref{eq:antecedent_score}) and cluster score $f_c$.
To avoid overlapping between $f_a$ and $f_c$, we set $f_c$ as $0$ if the cluster is the initial cluster (L6). 
Thus, $f_c$ becomes the consultation such that when $f_c > 0$, the span $g_x$ is likely to match the cluster $C_y$, and vice versa. 
$f_c$ is computed by \texttt{FFNN} similar to $f_a$, and $\phi(C_y)$ is the meta-feature such as the cluster size.

\begin{algorithm}
  \caption{Antecedent Ranking for \CM}\label{algo:cm}
  \begin{algorithmic}[1]
    \small
    \Procedure{ranking}{$g_1, \cdots, g_N$}
    \State $C_{i = 1, \cdots, N} \gets g_i$
    \State $R \gets $ ranking\_order($g_1, \cdots, g_N$)
    \For{$x = R_1 \cdots R_N$}
        \For{$ y \in \mathcal{Y}(x)$} \Comment{Parallelized}
            \State $f_c (g_x, C_y) \gets 0$ if $C_y = g_y$
            \State $s_x(y) \gets f_a (g_x, g_y) + f_c (g_x, C_y, \phi(C_y))$
        \EndFor
        \State $y' \gets \text{argmax}_{y \in \mathcal{Y}(x)} s_x(y)$
        \If{$y' \neq \epsilon$}
            \State merge $C_x$ and $C_{y'}$
        \EndIf
    \EndFor
    \State \textbf{return} $s_1, \cdots, s_N$
    \EndProcedure
  \end{algorithmic}
\end{algorithm}

\noindent Two simple configurations can be tuned for \texttt{CM}. We can have the sequential left-to-right ranking order or the easy-first order (L3) whose sequence is ordered by each span's max antecedent score, building the most confident clusters first \citep{ng-cardie-2002-improving,clark-manning-2016-improving}. There can be element-wise mean or max-reduction among the spans in the two merging clusters (L10).

Distinguished from \citet{wiseman-etal-2016-learning}, clusters in \texttt{CM} are searched and merged in training without the use of oracle clusters, closing the gap between training and test time.

\begin{table*}[htbp!]
\small
\centering
\begin{tabular}{>{\quad}cccccccccccccc}
\toprule
 & \multicolumn{3}{c}{MUC} & & \multicolumn{3}{c}{B\textsuperscript{3}} & & \multicolumn{3}{c}{CEAF\textsubscript{$\phi_4$}}\\
 \cmidrule{2-4} \cmidrule{6-8} \cmidrule{10-12}
 & P & R & F1 & & P & R & F1 & & P & R & F1 & Avg. F1 & Avg-M \\
\midrule
%\citet{wiseman-etal-2016-learning} & 77.5 & 69.8 & 73.4 & & 66.8 & 57.0 & 61.5 & & 62.1 & 53.9 & 57.7 & 64.2 \\
%\citet{clark-manning-2016-improving} & 79.2 & 70.4 & 74.6 & & 69.9 & 58.0 & 63.4 & & 63.5 & 55.5 & 59.2 & 65.7 \\
\tt L-17 & 78.4 & 73.4 & 75.8 & & 68.6 & 61.8 & 65.0 & & 62.7 & 59.0 & 60.8 & 67.2 & - \\
\tt L-18 & 81.4 & 79.5 & 80.4 & & 72.2 & 69.5 & 70.8 & & 68.2 & 67.1 & 67.6 & 73.0 & - \\
\tt F-19 & 85.4 & 77.9 & 81.4 & & 77.9 & 66.4 & 71.7 & & 70.6 & 66.3 & 68.4 & 73.8 & - \\
\tt K-19 & 82.6 & 84.1 & 83.4 & & 73.3 & 76.2 & 74.7 & & 72.4 & 71.1 & 71.8 & 76.6 & - \\
\tt J-19 & 84.7 & 82.4 & 83.5 & & 76.5 & 74.0 & 75.3 & & 74.1 & 69.8 & 71.9 & 76.9 & - \\
\tt J-20 & 85.8 & 84.8 & 85.3 & & 78.3 & 77.9 & 78.1 & & 76.4 & 74.2 & 75.3 & 79.6 & - \\
\midrule
\tt BERT & 85.0 & 82.5 & 83.8 & & 77.3 & 74.0 & 75.6 & & 74.9 & 70.7 & 72.8 & 77.4 & 77.3 ($\pm$0.1) \\
\tt SpanBERT & 85.7 & 85.3 & 85.5 & & 78.6 & 78.6 & 78.6 & & \bf 76.8 & 74.8 & 75.8 & 79.9 & 79.7 ($\pm$0.1) \\
+ \AAA & \bf 86.1 & 84.8 & 85.4 & & \bf 79.3 & 77.3 & 78.3 & & 76.0 & 74.7 & 75.4 & 79.7 & 79.4 ($\pm$0.2) \\
+ \EE  & 85.7 & 84.5 & 85.1 & & 78.5 & 77.4 & 77.9 & & 76.7 & 73.4 & 75.0 & 79.4 & 78.9 ($\pm$0.4) \\
+ \SC  & 85.5 & 85.2 & 85.4 & & 78.4 & 78.5 & 78.4 & & 76.5 & 74.1 & 75.2 & 79.7 & 79.2 ($\pm$0.3) \\
+ \CM  & 85.9 & \bf 85.5 & \bf 85.7 & & 79.0 & \bf 78.9 & \bf 79.0 & & 76.7 & \bf 75.2 & \bf 75.9 & \bf 80.2 & \textbf{79.9} ($\pm$0.2) \\
\bottomrule
\end{tabular}
\caption{Best results on the test set of the CoNLL'12 English shared task. The averaged F1 of MUC, B\textsuperscript{3}, CEAF\textsubscript{$\phi_4$} is the main evaluation metric. Avg-M: the mean Avg-F1 and its standard deviation from five developments. The mean and stdev of other metrics are provided in Appendix~\ref{append:results}. See Figure~\ref{fig:coref-history} for acronyms of the previous works.}
\label{table:results}
\end{table*}

%L-17: \citet{lee-etal-2017-end}
%L-18: \citet{lee-etal-2018-higher}
%F-19: \citet{fei-etal-2019-end}
%K-19: \citet{kantor-globerson-2019-coreference}
%J-19: \citet{joshi-etal-2019-bert}
%J-20: \citet{spanbert-joshi}

\section{Experiments}
\label{sec:exp}

For our experiments, the CoNLL 2012 English shared task dataset is used \citep{pradhan-etal-2012-conll}.
Given the end-to-end coreference system in Section~\ref{subsec:baseline}, six models are developed as follows:\footnote{Appdendix~\ref{append:experiments} provides details of our experimental settings.} 

\begin{itemize}
\small\setlength\itemsep{0em}
\item \texttt{BERT}: BERT \cite{devlin-etal-2019-bert} as the encoder
\item \texttt{SpanBERT}: SpanBERT \cite{spanbert-joshi} as the encoder
\item \texttt{+AA}: \texttt{SpanBERT} with attended antecedent (\textsec{subsec:previous})
\item \texttt{+EE}: \texttt{SpanBERT} with entity equalization (\textsec{subsec:previous})
\item \texttt{+SC}: \texttt{SpanBERT} with span clustering (\textsec{subsec:new})
\item \texttt{+CM}: \texttt{SpanBERT} with cluster merging (\textsec{subsec:new})
\end{itemize}

\noindent Note that \noindent \texttt{BERT} and \texttt{SpanBERT} completely rely on only local decisions without any HOI. 
Particularly, \texttt{+AA} is equivalent to \citet{spanbert-joshi}.

\subsection{Results}
\label{subsec:results}

Table~\ref{table:results} shows the best results in comparison to previous state-of-the-art systems. We also report the mean scores and standard deviations from 5 repeated developments, which we could not find from the previous works.
%Each model is developed 5 times and the max and mean scores are reported, 
%Since deep learning models are often not reproducible, it is important to report the mean scores with standard deviations to ensure the robustness, which we could not find from any of the previous works.
%We hope the experimental design suggested here will encourage researchers working on this task to evaluate the robustness of their models, instead of just reporting the single best scores.

The impact of SpanBERT over BERT is clear, showing 2.4\% improvement on average.
However, none of the HOI models shows a clear advantage over \texttt{SpanBERT} which adapts no HOI.
In fact, all HOI models except for \CM\ show negative impact.
The best result is achieved by \CM\ with the Avg-F1 of 80.2, surpassing the previous best result of 79.6 based on \textit{c2f-coref} reported by \citet{spanbert-joshi}.

%Our baseline model only uses local decisions (Sectoin~\ref{subsec:baseline}). Four HOI approaches \texttt{AA}, \texttt{EE}, \texttt{SC}, \texttt{CM} are experimented on the same baseline model (Section~\ref{subsec:previous}, \ref{subsec:new}). 
%Note that the baseline with SpanBERT + \texttt{AA} is equivalent to \citet{spanbert-joshi}. 
%Note that our baseline with SpanBERT\textsubscript{Large} adding \texttt{AA} is equivalent to \citet{spanbert-joshi}. 
%The model with \texttt{CM} \footnote{The best configuration is sequential order and max reduction.} achieves the best CoNLL score 80.2, out-performing the previous state-of-the-art by 0.6\%.

%It should be mentioned that the reported scores in Table~\ref{table:results} are from the best models out of 5 repeated experiments of each approach. However, that might not reflect the true performance due to variance, and it is unclear how previous work selects the single model to report. Therefore, we also report the macro-average of each metric in appendices, and we encourage future work also provides macro-average to better reflect the performance.

%It is surprising that the baseline model with only local decisions out-performs all HOI approaches except for \texttt{CM}. It indicates the score from \citet{joshi-etal-2019-bert} might be under-reported and could have been slightly higher by just removing the higher-order inference part.

\subsection{Impact Analysis of HOI}
\label{subsec:impact}

Three HOI methods based on span refinement, \texttt{AA}, \texttt{EE}, and \texttt{SC}, show negative impact upon local decisions. 
We suspect that error propagation from antecedent-ranking may downgrade the quality of refinement. 
On the other hand, \texttt{CM} shows marginal improvement, suggesting that maintaining entity clusters can be superior to span refinement, at the cost of more inference time from the sequential ranking process.
To analyze the direct impact of HOI, we take the trained models of each HOI method and evaluate them on the test set while turning off HOI, making it compatible to \texttt{SpanBERT}.

The averaged performance drop w.r.t Avg-F1 after turning off HOI is less than 0.2 for all methods\LN (Appendix~\ref{append:analysis}), implying that none of the HOI method has a significantly direct impact to the final performance of the model using SpanBERT.

\begin{table}[htbp!]
\small
\centering
\begin{tabular}{c|cccc}
%\toprule
\multicolumn{1}{c|}{} & \multicolumn{1}{c}{W2C} & \multicolumn{1}{c}{C2W} & \multicolumn{1}{c}{C2C} & \multicolumn{1}{c}{W2W} \\
\midrule
$\;$ + \texttt{AA} & 240.8 (1.3) & 241.2 (1.3) & 16262.2 & 2168.4 \\
$\;$ + \texttt{EE} & 244.1 (1.3) & 245.3 (1.3) & 16183.3 & 2136.3 \\
$\;$ + \texttt{SC} & 248.2 (1.3) & 262.0 (1.4) & 16184.4 & 2146.0 \\
$\;$ + \texttt{CM} & 226.4 (\textbf{1.2}) & 235.0 (\textbf{1.2}) & 16446.0 & 2180.0 \\
%\bottomrule
\end{tabular}
\caption{Averaged statistics on the test set prediction of four HOI approaches. W2C represents the number of mentions that are linked to a \textbf{W}rong antecedent before HOI and are linked to a \textbf{C}orrect antecedent after HOI; vice versa for C2W. C2C/W2W is the number of mentions that are both linked to \textbf{C}orrect/\textbf{W}rong antecedents before and after HOI. Parentheses indicate the percentage of corresponding numbers per row.}
\label{table:link_change}
\vspace{-3ex}
\end{table}

In further investigation, we examine the change of coreferent links w.r.t their correctness. Specifically, Table~\ref{table:link_change} shows the four types of link changes before and after HOI. It demonstrates that the benefits from HOI is diminished because the effects are two-sided: there are roughly same amounts of links (about 1\%) becoming correct or wrong after HOI, therefore neither HOI method leads to much improvement overall.

It is worth mentioning that the impact of HOI is not limited to only global decisions.
HOI implicitly\LN serves as a way of regularization that impacts local decisions as well, since HOI and local ranking are mutually dependent during training.
Such indirect influence of HOI makes it difficult to assess its true impact, which we will explore more in the future.
%Error propagation between HOI and local ranking can lead to either better or worse performance; 

%Furthermore, there is a discrepancy between the HOI impact and the final performance: the impact after turning off \texttt{SC} is 0.15, higher than other HOI approaches, while \texttt{SC} has the lowest CoNLL score shown in Table~\ref{table:results}. 

\subsection{Analysis of Pronoun Resolution}
\label{subsec:pronoun}

\begin{table}[htbp!]
\small
\centering
\begin{tabular}{c|cc|cc|c}
%\toprule
\multicolumn{1}{c|}{} & \multicolumn{1}{c}{SP} & \multicolumn{1}{c|}{PS} & \multicolumn{1}{c}{FL} & \multicolumn{1}{c|}{WL} & \multicolumn{1}{c}{BC} \\
\midrule
\tt BERT     & 2.3 & 6.5 & 213.8 & 186.3 & 48.8 (3.5) \\
\tt SpanBERT & 2.8 & 6.6 & 218.3 & 168.0 & \textbf{43.8} (2.7) \\
$\;$ + \texttt{AA} & \bf 1.8 & 8.8 & 214.2 & \bf 159.4 & 44.8 (2.4) \\
$\;$ + \texttt{EE} & \bf 1.8 & \bf 5.5 & 210.0 & 165.3 & 44.0 (2.5) \\
$\;$ + \texttt{SC} & 3.8 & 7.2 & 223.6 & 170.0 & 45.4 (3.0) \\
$\;$ + \texttt{CM} & 3.0 & 6.6 & \bf 208.0 & 162.2 &\textbf{43.8} (2.6) \\
%\bottomrule
\end{tabular}
\caption{Averaged statistics on the test set prediction of different approaches. SP is the number of coreferent links from \textbf{S}ingular to \textbf{P}lural personal pronouns; vice versa for PS. FL (False Link) and WL (Wrong Link) is the number of conreferent link errors that involve two personal pronouns. BC is the number of clusters that contain both singular and plural pronouns, and the parentheses indicate the numbers of BC that contain ambiguous pronouns such as ``you''.}
\label{table:analysis}
\vspace{-3ex}
\end{table}

\paragraph{Direct Inference}

For the error analysis, we examine the direct inference between two personal pronouns.\footnote{Ambiguous pronouns such as ``you'' are excluded in direct inference analysis, and included in indirect inference analysis.} 
SP/PS in Table~\ref{table:analysis} shows the numbers of links that one pronoun incorrectly selects another pronoun with different plurality as its antecedent. 
We find that adapting HOI shows slightly higher impact than switching to a more advanced encoder. 
\texttt{AA} can reinforce the pronoun representation to bias towards singularity and lead to lower SP error and higher PS error, while the difference between \texttt{BERT} and \texttt{SpanBERT} is trivial on SP/PS.

\noindent We also look at the general types of coreferent errors involving two pronouns. 
False Link (FL) falsely links a non-anaphoric pronoun to another pronoun as antecedent; Wrong Link (WL) links an anaphoric pronoun to another wrong pronoun as antecedent. Table~\ref{table:analysis} shows that \texttt{EE} and \texttt{CM} reduce FL errors by 4+\%, suggesting that the aggregation of non-local features indeed leads to more conservative linking decisions. 
However, adapting an advanced encoder shows higher impact on WL errors, as \texttt{SpanBERT} reduces almost 10\% compared to \texttt{BERT}, implying that representation learning is still more important for semantic matching.

\paragraph{Indirect Inference}

The plurality of ambiguous pronouns such as \textit{you} depends on the context. Two indirect links of (\textit{he}, \textit{you}) and (\textit{you}, \textit{they}) can be common to induce incorrect clusters that contain both singular and plural pronouns \citep{wiseman-etal-2016-learning,lee-etal-2018-higher}. Table~\ref{table:analysis} shows the numbers of these erroneous clusters in prediction. Surprisingly, very few of these clusters contain ambiguous pronouns in either approach. % (fewer than 3 in most approaches). 
This observation moderates the long-standing movitation of HOI.

Additionally, the change of representation from BERT to SpanBERT has far more impact that reduces 10\% of these erroneous clusters, while the four HOI methods fail to show significant difference compared to \texttt{SpanBERT}.

\section{Conclusion}
\label{sec:conclusion}

We implement the end-to-end coreference resolution model and investigate four higher-order inference methods, including two of our own methods. 
Our best model shows the new result of 80.2 on the CoNLL 2012 dataset. 
We thoroughly analyze the empirical effectiveness of HOI and demonstrate why it fails to boost performance on the CoNLL 2012 dataset compared to the improvement from encoders. We show that current HOI does not meet up with the original motivation, suggesting that a new perspective of HOI is needed for this task in the era of deep learning-based NLP.
\section*{Acknowledgments}

We gratefully acknowledge the support of the AWS Machine Learning Research Awards (MLRA).
Any contents in this material are those of the authors and do not necessarily reflect the views of AWS.

\bibliography{emnlp2020}
\bibliographystyle{acl_natbib}

\cleardoublepage\appendix
\begin{table*}[hb]
\small
\centering
\begin{tabular}{>{\quad}cccccccc}
\toprule
 & \multicolumn{1}{c}{MUC} & & \multicolumn{1}{c}{B\textsuperscript{3}} & & \multicolumn{1}{c}{CEAF\textsubscript{$\phi_4$}}\\
 \cmidrule{2-2} \cmidrule{4-4} \cmidrule{6-6}  
 & F1 & & F1 & & F1 & & Avg. F1\\
\midrule
\tt BERT & 83.7 ($\pm$ 0.1) & & 75.5 ($\pm$ 0.1) & & 72.6 ($\pm$ 0.1) && 77.3 ($\pm$ 0.1) \\
\tt SpanBERT & 85.3 ($\pm$ 0.1) & & 78.4 ($\pm$ 0.1) & & 75.5 ($\pm$ 0.3) && 79.7 ($\pm$ 0.1) \\
+ \AAA & 85.2 ($\pm$ 0.2) & & 78.1 ($\pm$ 0.2) & & 75.0 ($\pm$ 0.2) && 79.4 ($\pm$ 0.2) \\
+ \EE & 85.0 ($\pm$ 0.1) & & 77.7 ($\pm$ 0.2) & & 74.7 ($\pm$ 0.2) && 78.9 ($\pm$ 0.4) \\
+ \SC & 85.1 ($\pm$ 0.2) & & 77.9 ($\pm$ 0.3) & & 74.7 ($\pm$ 0.3) && 79.2 ($\pm$ 0.3) \\
+ \CM & \textbf{85.5} ($\pm$ 0.2) & & \textbf{78.5} ($\pm$ 0.3) & & \textbf{75.6} ($\pm$ 0.2) && \textbf{79.9} ($\pm$ 0.2) \\
\bottomrule
\end{tabular}
\caption{Results on the test set of the CoNLL'12 English shared task data. Macro-average is reported for each F1 score from 5 repeated developments of each approach. See Section~\ref{sec:exp} for the approaches.}
\label{table:results_avg}
\end{table*}

\section{Appendices}
\label{sec:appendix}

\subsection{Experimental Settings}
\label{append:experiments}

We implement the experimented models using PyTorch. BERT\textsubscript{Large} and SpanBERT\textsubscript{Large} are used as encoders. For each experiment, the best performed model on the development set is selected and evaluated on the test set.

\paragraph{Hyperparameters and Implementation}
Similar to \citet{joshi-etal-2019-bert,spanbert-joshi}, documents are split into independent segments with maximum 384 word pieces for BERT\textsubscript{Large} and 512 for SpanBERT\textsubscript{Large}. In our final setting, BERT-parameters and task-parameters have separate learning rates ($1 \times 10^{-5}$ and $3 \times 10^{-4}$ respectively), separate linear decay schedule, and separate weight decay rates ($10^{-2}$ and $0$ respectively). Models are trained 24 epochs with dropout rate 0.3.

The implementation of \texttt{EE} is based on the Tensorflow implementation from \citet{kantor-globerson-2019-coreference} which requires $\mathcal{O}(k^2)$ memory with $k$ being the number of extracted spans, while other HOI approaches only requires $\mathcal{O}(k)$ memory \footnote{The maximum number of antecedents for all models is set to 50 which is constant.}. To keep the GPU memory usage within 32GB, we limit the maximum number of span candidates for \texttt{EE} to be 300, which may have a negative impact on the performance.

Experiments are conducted on Nvidia Tesla V100 GPUs with 32GB memory. The average training time is around 7 hours for \texttt{BERT} and \texttt{SpanBERT} without HOI, and ranges from 9 - 15 hours with HOI methods.

\subsection{Results}
\label{append:results}

Table~\ref{table:results_avg} reports the macro-average F1 scores out of 5 repeated developments of each approach. \texttt{CM} still has the best performance with 79.9 averaged F1 score. Span refinement-based HOI approaches, \texttt{AA}, \texttt{EE}, and \texttt{SC}, still have lower F1 scores than the local-only \texttt{SpanBERT}.

We do not find different configurations for \texttt{CM} make any huge impact to the performance. The final configuration for \texttt{CM} is sequential order and max reduction (Algorithm~\ref{algo:cm}).

\subsection{Analysis}
\label{append:analysis}

\begin{table}[htbp!]
\centering
\begin{tabular}{c|c}
\toprule
\AAA & -0.02 ($\pm$ 0.06) \\
\EE & 0.03 ($\pm$ 0.07) \\
\SC & 0.11 ($\pm$ 0.10) \\
\CM & 0.04 ($\pm$ 0.04) \\
\bottomrule
\end{tabular}
\caption{Performance drop on CoNLL'12 English test set after turning off the corresponding HOI in trained models.}
\label{table:impact}
\end{table}

Table~\ref{table:impact} shows the averaged performance drop and its standard deviations w.r.t Avg-F1 after turning off the corresponding HOI in trained models, to see the direct performance impact of HOI over local decisions.

\paragraph{Pronoun Resolution}

In our analysis, the following personal pronouns are regarded as ambiguous pronouns: ``you'', ``your'', ``yours''.

\end{document}